\documentclass[letterpaper, 10 pt, conference]{ieeeconf}  

\IEEEoverridecommandlockouts                              

\usepackage{booktabs}    
\usepackage{xcolor}      
\usepackage{graphicx}    
\usepackage{colortbl}    
\usepackage{subcaption}  

\definecolor{lightgray}{gray}{0.9}  
\definecolor{gray20}{gray}{0.8}  
\definecolor{lightblue}{RGB}{210, 230, 255}

\def\checkmark{\tikz\fill[scale=0.4](0,.35) -- (.25,0) -- (1,.7) -- (.25,.15) -- cycle;}

\usepackage{graphicx}

\usepackage[]{xcolor}
\usepackage{todonotes}
\usepackage{bm}
\usepackage{amsmath}
\usepackage{booktabs}
\usepackage{multirow}
\usepackage{subcaption}
\usepackage{graphicx}
\usepackage{pdfpages}

\usepackage{algorithm}
\usepackage{listings}

\usepackage{hyperref}

\title{\LARGE \bf
FLAME: A Federated Learning Benchmark for Robotic Manipulation
}

\author{Santiago Bou Betran$^{*,1}$, Alberta Longhini$^{*,1}$, Miguel Vasco$^{1}$, Yuchong Zhang$^{1}$ and Danica Kragic$^{1}$
\thanks{*Equal Contribution, $^{1}$All authors are with KTH Royal Institute of Technology, Stockholm, Sweden {\tt\small {sbb, albertal, miguelsv, yuchongz, dani}@kth.se}}}

\begin{document}

\maketitle
\thispagestyle{empty}
\pagestyle{empty}

\begin{abstract}
Recent progress in robotic manipulation has been fueled by large-scale datasets collected across diverse environments. Training robotic manipulation policies on these datasets is traditionally performed in a centralized manner, raising concerns regarding scalability, adaptability, and data privacy. While federated learning enables decentralized, privacy-preserving training, its application to robotic manipulation remains largely unexplored. We introduce FLAME (Federated Learning Across Manipulation Environments), the first benchmark designed for federated learning in robotic manipulation. FLAME consists of: (i) a set of large-scale datasets of over 160,000 expert demonstrations of multiple manipulation tasks, collected across a wide range of simulated environments; (ii) a training and evaluation framework for robotic policy learning in a federated setting. We evaluate standard federated learning algorithms in FLAME, showing their potential for distributed policy learning and highlighting key challenges. Our benchmark establishes a foundation for scalable, adaptive, and privacy-aware robotic learning. The code is publicly available at \url{https://github.com/KTH-RPL/ELSA-Robotics-Challenge}.

\end{abstract}

\section{INTRODUCTION}
\label{sec:introduction}

Recent advances in robot learning have significantly enhanced the autonomy and capability of robotic manipulation systems~\cite{zhao2024aloha,black2024pi_0,brohan2023rt}. Large-scale datasets of interaction data collected across different institutions have played a crucial role in enabling robots to learn from extensive demonstrations and refine their skills through imitation and reinforcement learning~\cite{khazatsky2024droid,fang2023rh20t,o2024open,walke2023bridgedata}. However, as these datasets grow with data coming from real-world applications such as domestic environments, data privacy and decentralized learning become critical concerns~\cite{qiu2019guest,yankson2021empirical}. In this context, training manipulation policies in a distributed, privacy-preserving manner remains largely unexplored.

\begin{figure}[]
    \centering
    \includegraphics[width=0.98\linewidth]
    {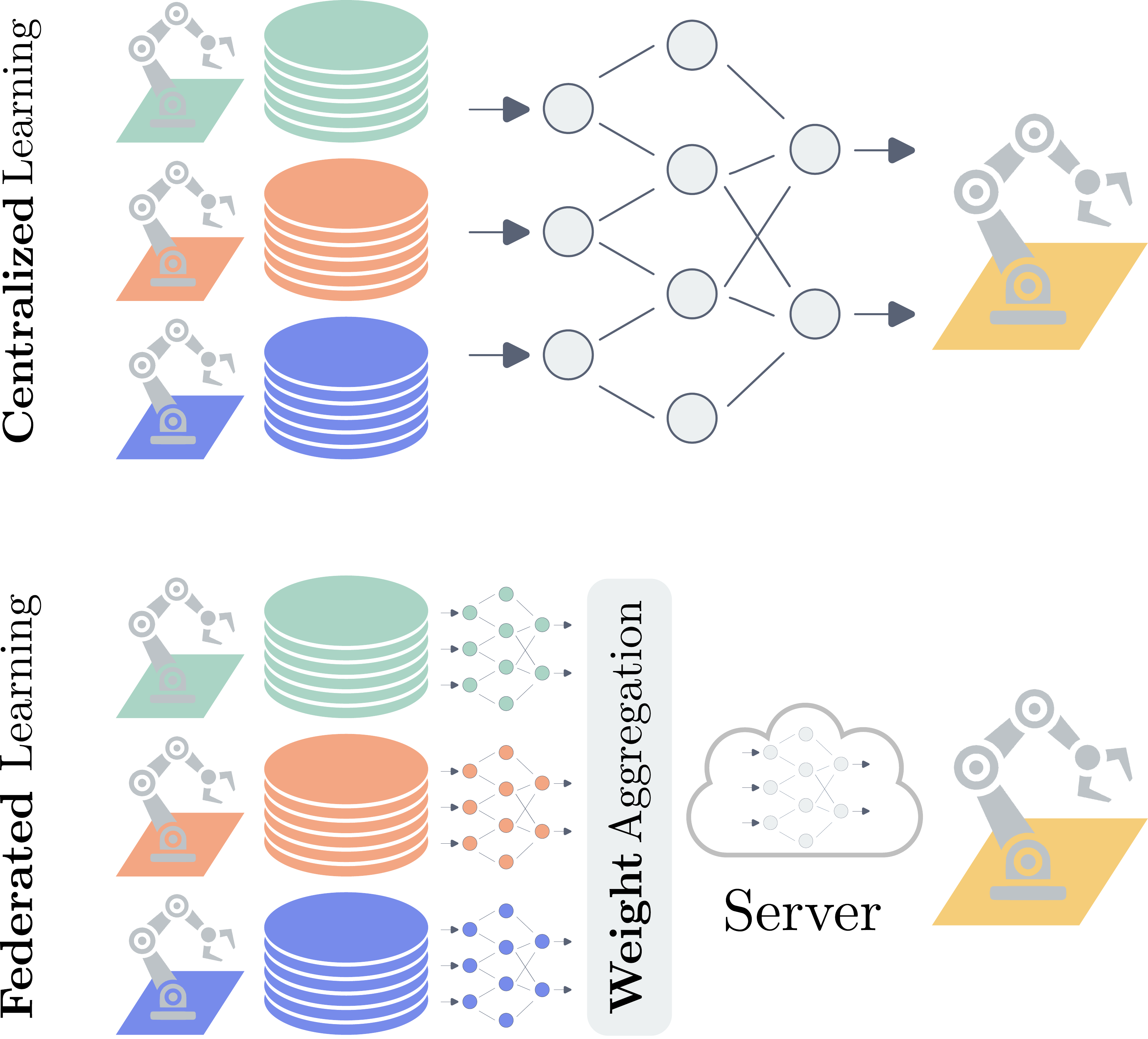}
    \caption{\textbf{Learning manipulation policies through federated learning.} In centralized training (top), a global model is trained iteratively over a large dataset stored in a central server. In contrast, federated learning (bottom) enables distributed robots to locally train models on their own data and periodically share updates with a central server, preserving data privacy. In this work, we contribute FLAME, the first benchmark of federated learning for robotic manipulation.}
    \vspace{-3ex}
    \label{fig:first_image}
\end{figure}

Traditional methods for learning manipulation policies from large-scale datasets rely on centralized training~\cite{zhao2024aloha,brohan2023rt,driess2023palm,kim2024openvla}, where a behavioral model is trained iteratively over an entire dataset stored in a centralized manner (illustrated in Fig.\ref{fig:first_image}, top). However, we envision a paradigm shift where robots continuously acquire knowledge from their own experiences in situated, diverse, real-world environments and collaboratively update a global policy model, while keeping their experience data local and private (Fig.\ref{fig:first_image}, bottom). Such an approach, termed federated learning (FL)~\cite{mcmahan2017communication}, allows distributed robots to contribute to a shared model without centralizing sensitive data. Instead of learning from static datasets, robots could dynamically improve their manipulation policies by leveraging experiences across multiple settings, leading to more adaptive and resilient behaviors. While federated learning has gained traction in robotics, particularly in mobile robot navigation~\cite{casado2022federated,yu2022towards} and object grasping~\cite{kang2023fogl}, there remains a notable gap: no standardized benchmarks exist to evaluate FL frameworks specifically for robotic systems, and robotic manipulation in particular has yet to be explored through the lens of federated learning.

To address this gap, we introduce FLAME (\emph{Federated Learning Across Manipulation Environments}), a benchmark designed to evaluate federated learning strategies for robotic manipulation tasks. Our benchmark provides a set of large-scale datasets of diverse manipulation tasks collected across multiple settings, incorporating variations in lighting conditions, textures, object appearances, and camera viewpoints, encompassing over $15\text{M}$ data samples. Furthermore, FLAME integrates these distributed datasets into a FL framework, enabling a rigorous evaluation of FL algorithms and serving as a foundation for developing more robust federated learning manipulation strategies.

We employ our proposed benchmark to evaluate different federated learning baselines across a diverse set of environments and four different manipulation tasks. Our results highlight the potential of federated learning as a scalable and privacy-preserving strategy for robotic manipulation, paving the way for real-world deployment and the continual learning of adaptive robotic systems.

In summary, our contributions are as follows: (i) FLAME, a federated learning benchmark for robotic manipulation, consisting of a large-scale dataset collected across multiple simulated environments and tasks, and a federated learning and evaluation framework designed for robotic manipulation; (ii) a rigorous evaluation of different federated learning baselines, demonstrating the need for more research for federated approaches for robotic manipulation training in a distributed setting; (iii) to the best of our knowledge, this work presents the first application of federated learning to robotic manipulation strategies

We believe our benchmark will serve as a valuable resource for the robotics and machine learning communities, fostering advancements in robotic manipulation. The code and datasets of FLAME are publicly available at \url{https://github.com/KTH-RPL/ELSA-Robotics-Challenge}.


\section{Related Work}
\label{sec:rw}

\subsection{Federated Learning Benchmarks} Federated Learning is a machine learning paradigm that advocates for the distributed training of local models to learn a global model by aggregating locally-computed updates~\cite{mcmahan2017communication}. The best aggregation strategy is an active area of research~\cite{louizos2021federated,reddi2020adaptive,hsu2019measuring,blanchard2017byzantine}. The emergence of federated learning has spurred the creation of diverse benchmarks such as FedScale~\cite{lai2022fedscale}, FLBench~\cite{liang2021flbench}, FedLLM-Bench~\cite{ye2025fedllm}, and LEAF~\cite{caldas2018leaf} to evaluate FL methods on text, image, speech, or preference data. Despite covering different learning tasks, such as text and image classification~\cite{caldas2018leaf,liang2021flbench}, medical analysis~\cite{song2022flair}, and instruction tuning~\cite{ye2025fedllm}, these benchmarks focus exclusively on domains outside robotics. Consequently, there is still no established federated dataset or experimental protocol tailored specifically to robotic manipulation, highlighting the need for domain-specific solutions.

\subsection{Federated Learning for Robotics} In robotics, FL-based learning offers significant advantages by enabling the continuous, distributed training of models while preserving data privacy and leveraging edge computing to mitigate bandwidth limitations~\cite{xianjia2021federated,yu2022towards}. However, the application of federated learning in robotics remains relatively unexplored. Recent works on distributed reinforcement learning~\cite{na2023federated} have demonstrated its potential for achieving high-quality policy transfer while ensuring both data and model privacy, yet without explicitly addressing robotics tasks~\cite{qi2021federated}. In contrast, FL for robot navigation and obstacle avoidance has been investigated in~\cite{na2023federated, majcherczyk2021flow}, highlighting the benefits of multi-robot federated learning with continuous data collection. Additionally,~\cite{ho2022federated} applies FL to task scheduling for heterogeneous agents in a warehouse setting.  In the context of robotic manipulation,~\cite{kang2023fogl} introduces an FL-based algorithm for multi-robot grasping, demonstrating effective training even under heterogeneous and non-transferable data conditions. Despite these advances, FL for learning complete manipulation policies remains largely unexplored. Our work bridges this gap by introducing FL benchmarks that integrate local demonstrations while providing a global model for imitation-based manipulation tasks. Furthermore, we evaluate federated learning strategies for aggregating local models into a unified global policy, laying the foundation for federated learning in robotic manipulation.

\subsection{Benchmarks and Datasets for Robotic Manipulation}  Datasets and benchmarks in robotic manipulation increasingly emphasize standardized tasks, diverse object sets, and domain shifts, such as RoboNet~\cite{dasari2019robonet}, BridgeData~\cite{walke2023bridgedata}, or The Colosseum~\cite{pumacay2024colosseum}, particularly with the evolution of imitation learning and, more specifically, behavior cloning (BC). These efforts test how well policies adapt to task variations~\cite{zhu2020robosuite,luo2023fmb} and environment variations, such as lighting conditions, texture, or camera poses~\cite{yuan2023rl}. However, existing benchmarks generally rely on centralized data aggregation and do not incorporate the distributed data constraints inherent to FL scenarios.  Our work builds upon Colosseum by introducing a federation-compatible dataset and codebase specifically designed for FL-based imitation learning in robotic manipulation. By supporting local training across diverse environments and aggregating model updates in a privacy-preserving manner, we establish the first benchmark to evaluate federated learning in robotic manipulation tasks.

\section{FLAME}

\begin{figure*}[ht!]
    \centering
    \includegraphics[width=\linewidth]{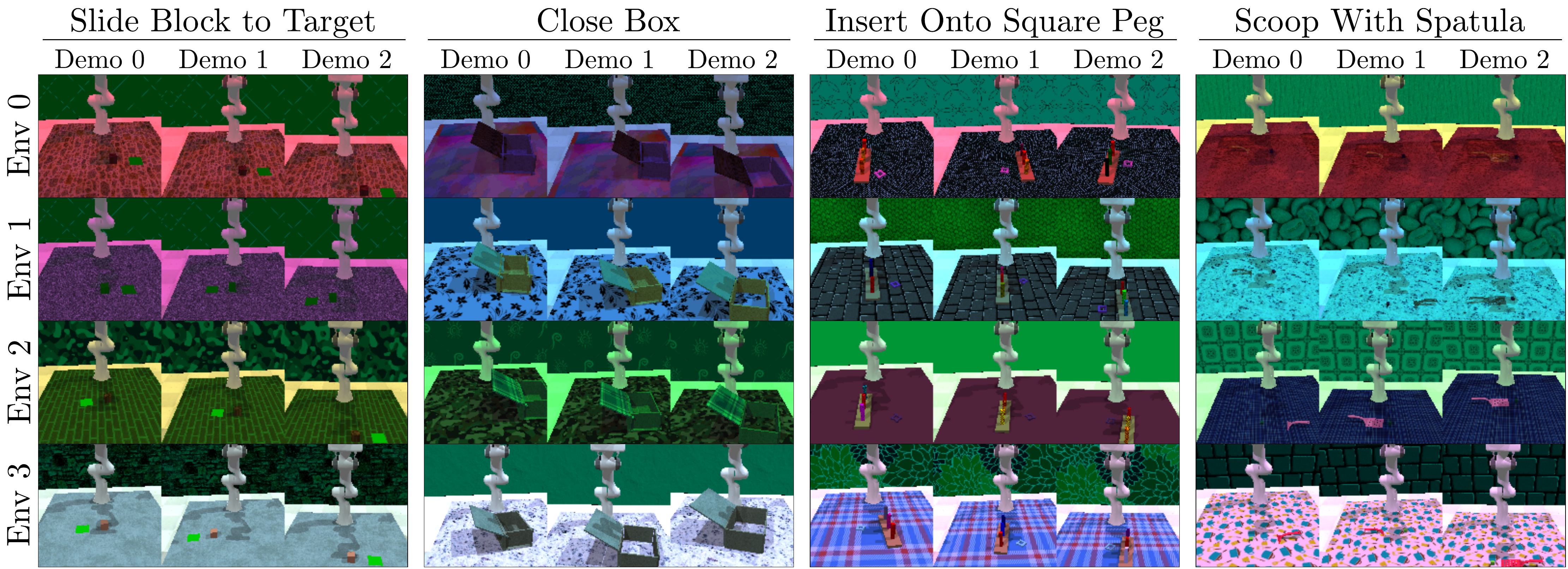}
    \caption{\textbf{Tasks, demonstrations and variations in FLAME}. The figure showcases four robotic manipulation tasks: \textit{Slide Block to Target}, \textit{Close Box}, \textit{Insert Onto Square Peg}, and \textit{Scoop With Spatula}. Each row represents a different environment instance (Env 0–3), introducing variations in background, object textures, lighting, and camera perspectives. Within each environment, three demonstrations (Demo 0–2) illustrate different executions of the same task, capturing the diversity in data collection used for training and evaluation in our federated learning framework. 
    }
    \label{fig:variations-example}
    \vspace{-3ex}
\end{figure*}

\label{sec:FLAME}

FLAME is a federated learning benchmark designed to train and evaluate robotic manipulation tasks under diverse environmental conditions in a distributed setting. Built upon RLBench~\cite{james2020rlbench}, it extends the Colosseum~\cite{pumacay2024colosseum} benchmark, consisting of $20$ diverse robotic manipulation tasks, each enabled with $14$ different perturbation factors to increase environment diversity. These perturbations, which include variations in object color, lighting conditions, and other scene properties, increase the complexity of the benchmark, leading to the instantiation of $20,371$ unique environments.

A key methodological difference between FLAME and~\cite{pumacay2024colosseum} lies in how these variations are treated during data collection. While~\cite{pumacay2024colosseum} considers each perturbed instance as a distinct task and collects only one demonstration per randomized environment, FLAME treats them as unique environments within a federated setting. This allows for the collection of a more extensive set of demonstrations per environment, with randomized initial conditions, ensuring that federated learning algorithms are exposed to a broad variability of environments. Additionally, FLAME introduces a structured indexing method that records the specific perturbation factors associated with each environment. This ensures that environments can be re-instantiated with identical conditions when needed for distributed learning, providing greater control over the evaluation process.

Another major advancement of FLAME is its integration within a federated learning framework, which enables systematic benchmarking of FL algorithms for robotic manipulation tasks. Unlike~\cite{pumacay2024colosseum}, which does not natively support federated learning, FLAME is designed to facilitate large-scale distributed training and evaluation. The combination of extensive environment diversity, structured data collection, and federated learning integration makes FLAME a powerful benchmark for investigating generalization and robustness in distributed robotic learning systems. In the remainder of this section, we provide further details into the environment definition and the perturbation factors, the methodology of data collection and environment indexing, and the FL integration for algorithm implementation and benchmarking.

\subsection{Environments and Variations}
Formally, in FLAME we define an environment ${E = \{Q, V\}}$ as the combination of one of the RLBench tasks $Q$ with a predefined set of factors of variations $V$. We highlight in Figure~\ref{fig:variations-example} examples of different tasks and factors of variation, such as predefined lightning conditions, background color, and camera position.  These perturbations can be categorized into five main types~\cite{pumacay2024colosseum}: (i) \emph{color variations} are applied to manipulated objects, scene illuminators, and tables, with RGB values sampled uniformly within a predefined range; (ii) \emph{texture modifications} involve randomly selecting from a set of 213 textures for objects, tables, and backgrounds, introducing additional visual diversity; (iii) \textit{object variations} include the placement of distractor objects, chosen from a fixed set of 78 unique 3D models, ensuring that learned policies remain robust to scene clutter; and (iv) \textit{physical properties}, such as friction and mass, are adjusted to modify the dynamics of object interactions, impacting grasping and manipulation strategies; (v) \emph{view variations} introduce small perturbations in angle and position of the camera to encourage generalization across different viewpoints. These factors collectively create a diverse and challenging dataset, facilitating the evaluation of federated learning algorithms in robotic manipulation under a wide range of domain shifts.




\subsection{Dataset Structure}



The large-scale nature of our federated training setup requires an adapted data-saving scheme that maximizes performance and ensures the safety and independence of each client's dataset. To this end, we design our training such that each client will be trained in a single unique environment, with a specified and replicable set of variation parameters. To do so, we sample a database of environments, encoded as a \texttt{JSON} configuration file, each collected environment is stored in this database with a unique client ID. 

We define the structure of the dataset based on the following hierarchy definition:

\begin{itemize}
    \item \emph{Environment}: At a top level, we define a set of $N$ environments, each consisting of the manipulation task $Q$ and the factors of variation $V$, randomly sampled within feasible limits, and stored in the JSON file. Each environment corresponds to a unique local agent for our federated learning setup.
    \item \emph{Episodes}: For each environment, we collect a predefined number of episodes $K$, using a pretrained, scripted, expert policy. At the beginning of each episode, we randomly sample the pose of the elements of interest in the scene, such as the effected object and the target. This ensures that the training process does not overfit to a specific range of manipulator movements.
\end{itemize}

\begin{figure*}[t]
    \centering
    \includegraphics[width=0.98\linewidth]{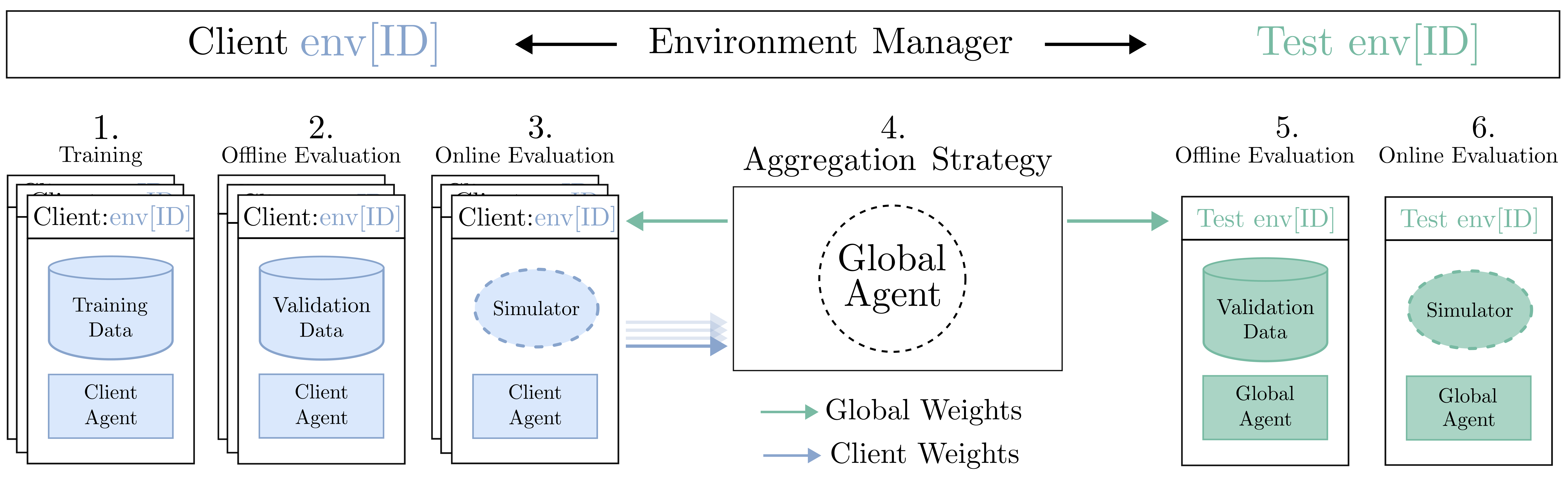}
    \caption{\textbf{The FLAME learning and evaluation framework}. Our federated learning framework starts by sampling a set of clients for local training (1). After training for a user-defined number of local epochs, we evaluate each local policy on offline validation data (2) and in online interactions within the simulator (3). After the evaluation, we aggregate the weights of the local policies using a predetermined federated learning method and send back the aggregated weights to initialize the local clients in the new round of training (4). We repeat steps (1-4) for a user-defined number of aggregation rounds. Following this, we select the best-performing global policy in the validation dataset and evaluate the final policy using the test dataset (5) and in the test simulator environments (6).}
    \label{fig:federated_learning}
    \vspace{-3ex}
\end{figure*}



\subsection{Federated Learning Framework}
To enable federated learning within our benchmark, we exploit FLOWER~\cite{beutel2020flower} as the backbone for distributed training across simulated environments. FLOWER is a Python library providing a scalable and flexible implementation of federated learning, supporting both homogeneous and heterogeneous client scenarios while providing distributed node computation capabilities. We built a customized wrapper around FLOWER to facilitate seamless interaction with the RLBench-based simulation, ensuring efficient data distribution, environment indexing, and federated model training.

Our wrapper extends FLOWER by introducing an environment management system that systematically indexes and assigns unique task variations to each of the federated clients. Each client environment configuration is determined by the sampled perturbation factors applied during data collection. This allows for distributed training across a diverse set of environments while maintaining consistency when re-instantiating a given environment. Additionally, we implement a structured data pipeline that handles the allocation of demonstrations, ensuring that each client receives a dataset that aligns with its assigned task variation.


Furthermore, we adapt FLOWER’s client-server architecture to accommodate scripted demonstrations as training data, as shown in Figure~\ref{fig:federated_learning}. Each client trains on a unique subset of the training dataset, evaluates its performance and transmits its parameter updates to the central aggregation server. The server, in turn, consolidates updates from multiple clients using a predefined aggregation algorithm (e.g., Federated Averaging~\cite{mcmahan2017communication}), allowing the global model to progressively generalize across varying task conditions. By leveraging FLOWER’s flexibility, we enable systematic benchmarking of federated learning algorithms within robotic manipulation tasks, assessing their performance under diverse environmental shifts and task variations. For a complete description of the training and evaluation procedure in FLAME please refer to Figure~\ref{fig:federated_learning}.



\section{EVALUATION}
\label{sec:evaluation}

\renewcommand{\arraystretch}{1.2} 
\setlength{\arrayrulewidth}{0.1pt} 

\begin{table*}[t]
\centering
\caption{\textbf{Factors of variation in the FLAME benchmark datasets}. The table presents the value ranges for each variation across the four manipulation tasks.}
\resizebox{0.9\textwidth}{!}{%
\begin{tabular}{lcccc}
\toprule
\rowcolor{lightgray} \textbf{Factor} & \textbf{Slide Block To Target} & \textbf{Close Box} & \textbf{Insert Onto Square Peg} & \textbf{Scoop With Spatula} \\ 
\midrule
Background Texture            & \checkmark & \checkmark & \checkmark & \checkmark \\
Camera Pose $(\Delta x/ \Delta y/ \Delta z)$ & (-0.05, 0.05) & (-0.05, 0.05) & (-0.05, 0.05) & (-0.05, 0.05) \\
Light Color (RGB)   & (0, 0.5) & (0, 0.5) & (0, 0.5) & (0, 0.5) \\
Object Color (RGB)   & (0, 1) & (0, 1) & (0, 1) & (0, 1) \\
Object Size       & – & 0.75 – 1.15 & 1.0 – 1.5 & 0.75 – 1.25 \\
Object Texture        & \checkmark & \checkmark & \checkmark & \checkmark \\
Table Color (RGB)   & (0.25, 1) & (0.25, 1) & (0.25, 1) & (0.25, 1) \\
Table Texture   & \checkmark & \checkmark & \checkmark & \checkmark \\

\bottomrule
\end{tabular}
}
\label{tab:flame_variations}
\end{table*}

\renewcommand{\arraystretch}{1.4} 
\setlength{\arrayrulewidth}{0.1pt} 
\begin{table*}[t]
\centering
\caption{\textbf{Performance of different federated learning methods in FLAME}. We show the RMSE and the normalized success rate of the different methods across the four manipulation tasks. The values are averaged across 10 test environments and 50 episodes for each task. The arrows indicate the direction of improvement. Best viewed with zoom.}
\label{tab:rmse_success}

\resizebox{\textwidth}{!}{%
\begin{tabular}{c c}

\subcaptionbox{Offline Evaluation: RMSE $\times 10^{-2} (\downarrow)$\label{tab:rmse}}{
    \begin{tabular}{lcccc}
    \toprule
    \rowcolor{lightgray} \textbf{Method} & \textbf{Slide Block} & \textbf{Close Box} & \textbf{Peg in Square} & \textbf{Scoop} \\
    \midrule
    \rowcolor{white} 
    \texttt{FedAvg}     & $2.64 \pm 0.13$  & $6.24 \pm 0.79$ & $30.99 \pm 0.52$ &$ 3.85\pm 0.21$ \\
    \texttt{FedAvgM}    & $13.45 \pm 0.26$  & $11.9 \pm 1.03$ & $38.58 \pm 0.38$ &$40.3 \pm 0.32$ \\
    \texttt{FedOpt}     & $2.76 \pm 0.13$  & $6.86 \pm 0.91$ & $30.75 \pm 0.50$ &$3.89 \pm 0.22$ \\
    \texttt{Krum}       & $7.24 \pm 2.23$  & $9.68 \pm 1.98$ & $28.13 \pm 7.06$ &$10.19 \pm 0.76$ \\
    \bottomrule
    \end{tabular}
}

& 
\subcaptionbox{Online Evaluation: Normalized Success Rate $(\uparrow)$\label{tab:success_rate}}{
    \begin{tabular}{lcccc}
    \toprule
    \rowcolor{lightgray} \textbf{Method} & \textbf{Slide Block} & \textbf{Close Box} & \textbf{Peg in Square} & \textbf{Scoop} \\
    \midrule
    \rowcolor{white} 
    \texttt{FedAvg}    & $0.24 \pm 0.43$  & $0.84 \pm 0.37$ & $0 \pm 0$ & $0 \pm 0$ \\
    \texttt{FedAvgM}   & $0 \pm 0$  & $0.54 \pm 0.49$ & $0 \pm 0$ & $0 \pm 0$ \\
    \texttt{FedOpt}    & $0.28 \pm 0.45$  & $0.70 \pm 0.46$ & $0 \pm 0$ & $0 \pm 0$ \\
    \texttt{Krum}      & $0.10 \pm 0.30$  & $0.86 \pm 0.38$ & $0 \pm 0$ & $0 \pm 0$ \\
    \bottomrule
    \end{tabular}
    }
\end{tabular}
}
\end{table*}

\subsection{Experimental Setup}
\label{sec:experimental}

\textbf{Tasks:} For our evaluation, we select four distinct tasks: $Q =\{$\textit{Slide block to target} (Slide Block), \textit{Close Box} (Close box),  \textit{Insert onto Square Peg} (Peg in Square), \textit{Scoop with Spatula} (Scoop)$\}$.  In each of these environments, the robot’s action space is composed of joint velocities as well as a binary command to open or close the gripper. Meanwhile, the observation space includes the robot’s joint positions and a single-view camera feed that supplies image observations. This setup allows us to capture both low-dimensional kinematic information and high-dimensional visual context. For each of these tasks, we follow the data collection procedure described in the previous section, containing $N=420$ different environments and $K=100$ demonstrations per environment, of which $400$ environments are used for training, $10$ for validation, and $10$ for testing. We highlight the selected factors of variation in Table~\ref{tab:flame_variations}.

\textbf{Local Imitation Learning Agent:} For each environment, we instantiate the local agent as a multi-modal neural network that integrates visual and low-dimensional state inputs to predict continuous control actions. The architecture employs a CNN encoder for processing RGB images of size (128, 128, 3), consisting of three convolutional layers with 32, 64, and 128 filters, respectively, each with a 3×3 kernel, stride of 2, and ReLU activation, followed by a fully connected layer of 256 units. In parallel, an MLP encoder processes the low-dimensional state input through two fully connected layers with 256 and 128 units, both using ReLU activations. The extracted visual and state features are concatenated into a 384-dimensional latent representation and passed through a policy network comprising two fully connected layers of 512 and 4 units, where the final layer applies a Tanh activation to constrain the action outputs to the range [-1, 1]. The model is trained using supervised learning with a Mean Squared Error (MSE) loss, optimizing the action predictions against expert demonstrations using the Adam optimizer with a learning rate of 1e-4. To enable decentralized training, the agent is deployed within a federated learning framework. To incorporate updates from multiple distributed environments, we benchmark different aggregation strategies, as described in the following section.

\textbf{Federated Learning Approaches:} We evaluate multiple federated learning baselines to assess their performance in decentralized robotic learning. In Federated Averaging~\cite{mcmahan2017communication} (\texttt{FedAvg}) local models are trained independently on client devices before being aggregated into a global model via weighted averaging. Federated average with Momentum~\cite{hsu2019measuring} (\texttt{FedAvgM}) extends this by incorporating a momentum-based update rule on the server side, which helps stabilize training and mitigate oscillations due to non-IID client data distributions. Krum~\cite{blanchard2017byzantine} (\texttt{Krum}) is an aggregation method designed for Byzantine-robust federated learning, selecting a single client update that minimizes the influence of potential adversarial models. Adaptive Federated Optimization~\cite{reddi2020adaptive} (\texttt{FedOpt}) generalizes FedAvg by enabling adaptive optimizer strategies, such as Adam, at the server, improving convergence in heterogeneous environments. 


\textbf{Training Setup:} We trained all federated learning models for a total of $30$ aggregation rounds, with each client performing $50$ local training epochs per round. The number of available clients was set to $400$, corresponding to the number of distinct training environments.  Due to computational constraints, $20$ random clients were selected per training round. Model validation and testing were conducted on $20$ previously unseen environments, with the best-performing model on the evaluation set subsequently assessed on the test set. All models were trained on a system equipped with a single NVIDIA A$100$ GPU, $64$ CPU cores, and $250$ GB of allocated memory.

\begin{figure*}[h!]
    \centering
    
    \begin{subfigure}[]{0.24\textwidth}
        \centering
        \includegraphics[width=1\textwidth]{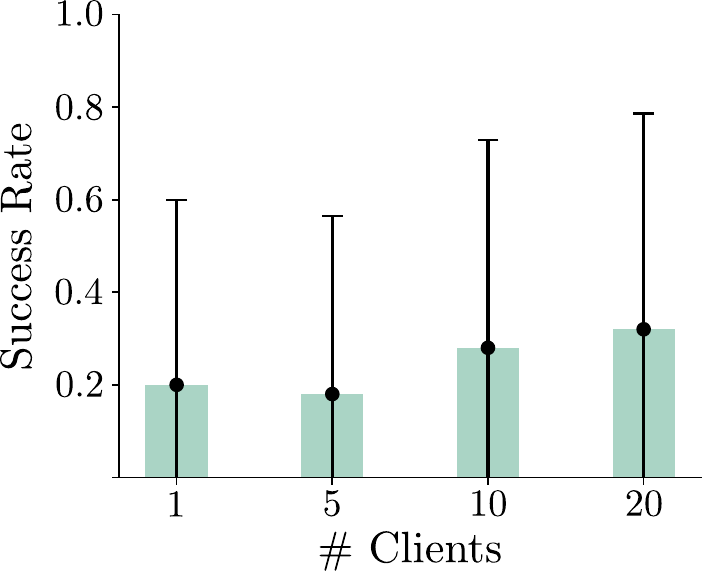}
        \caption{Local Clients}
        \label{fig:subfig1}
    \end{subfigure}
    \hfill
    \begin{subfigure}[]{0.24\textwidth}
        \centering
        \includegraphics[width=1\textwidth]{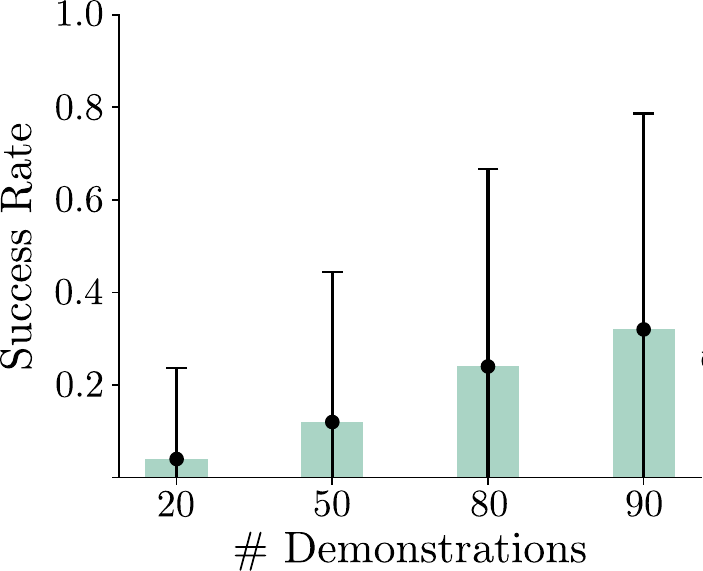}
        \caption{Demonstrations per Client}
        \label{fig:subfig2}
    \end{subfigure}
    \hfill
    \begin{subfigure}[]{0.24\textwidth}
        \centering
        \includegraphics[width=1\textwidth]{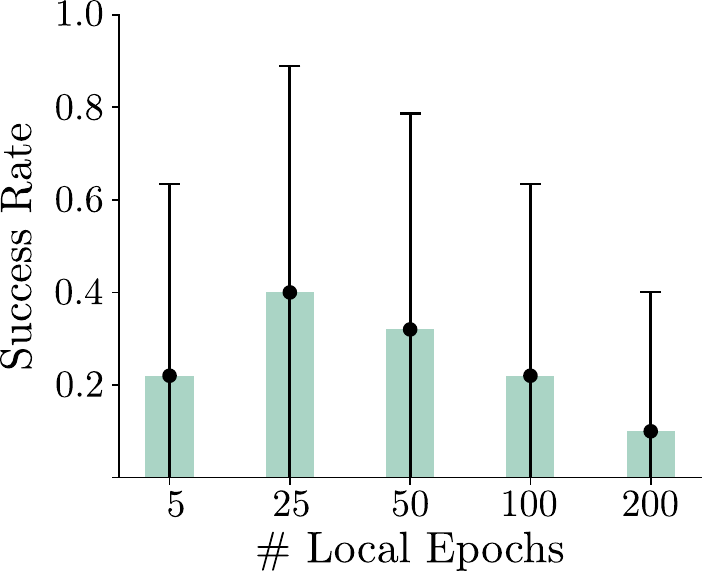}
        \caption{Local Training Epochs}
        \label{fig:subfig3}
    \end{subfigure}
    \hfill
    \begin{subfigure}[]{0.24\textwidth}
        \centering
        \includegraphics[width=1\textwidth]{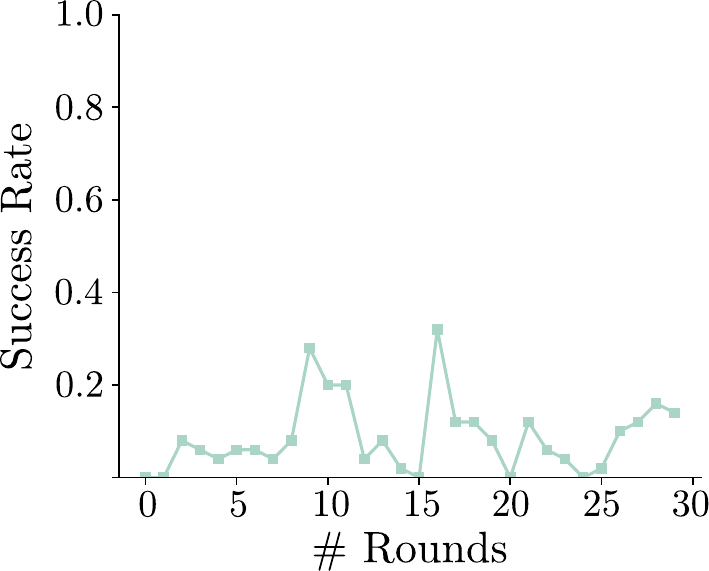}
        \caption{Aggregation Rounds}
        \label{fig:subfig4}
    \end{subfigure}

    \caption{\textbf{Ablation study of federated learning for robotic manipulation}. In this study, we consider the Slide Block task and the \texttt{FedAvg} method. We evaluate the performance of the final policy as a function of: (a) the number of local clients during training; (b) the number of demonstrations per client during training; (c) the number of local training epochs; (d) the number of aggregation rounds. All results are averaged over 50 episodes across 10 different test environments.}
    \vspace{-3ex}
    \label{fig:ablation_studies}
\end{figure*}


\subsection{Results}
We evaluate all federated learning approaches across the different tasks of the FLAME benchmark. We consider two different evaluation scenarios: an \emph{offline} evaluation scenario, in which we compute the RMSE between the predicted actions of the agent and the test actions; and an \emph{online} evaluation, where we employ the RLBench simulator and the variation configuration corresponding to the test environments to evaluate the success rate of the agent. We present our main results in Table~\ref{tab:rmse_success}.

\textbf{Variability across manipulation tasks:} The results highlight significant performance differences across the different tasks in FLAME. While most federated learning approaches are able to learn an effective policy in Close Box, they struggle to generalize in the Slide Block task and are unable to do so in both Peg in Square and Scoop. These differences are expected given the increased difficulty of these tasks: both Peg in Square and Scoop require precise manipulation of small objects, while Close Box and Slide Block only require a coarser manipulation strategy. This challenge is intentional: FLAME is designed to evaluate the generalization capabilities of current and future manipulation methods within a federated learning setup.

\textbf{Variability across FL methods:} The results also highlight significant performance differences across the different federated learning methods. The performance of each method is also highly dependent on the environment: for example, \texttt{Krum} achieves the highest average success rate in the Close Box task, yet struggles to perform in the Slide Block task in comparison with \texttt{FedAvg} and \texttt{FedAvgM}. This disparity raises two important observations. First, the need to evaluate novel federated learning methods across a wide variety of manipulation tasks to reliably estimate their performance, such as the ones available in FLAME. Secondly, the need to develop federated learning methods that are designed specifically for robotic manipulation tasks. We hope our results and benchmark inspire the community to advance federated learning methods tailored for robotic manipulation.

\textbf{Importance of complementary evaluations:} Finally, our results highlight the need to complement offline evaluations (traditionally employed in federated learning literature) with online evaluations to access the performance of these methods for robotic manipulation. While offline evaluation can provide insights into how these models predict actions compared to an expert policy, they do not always correlate with successful executions in the online setting: for example, \texttt{Krum} has a high RMSE in the Close Box task, yet achieves the highest success rate in the online evaluation. This underscores the importance of integrating simulation capabilities into federated learning frameworks, as done in FLAME, to better align offline evaluations with actual robot policy performance.

\subsection{Ablation Study}

We present an ablation study to understand how different hyperparameters of the federated learning setup influence the final performance of the robot policy. To do so, we consider the Slide Block task and the \texttt{FedAvg} method. We explore how: (i) the number of local clients during training; (ii) the number of training demonstrations per client; (iii) the number of local training epochs per client; and (iv) the number of central aggregation rounds influence the performance of the final agent. The remaining hyperparameters are set to the values described in Section~\ref{sec:experimental}. The results of our ablation study are presented in Figure~\ref{fig:ablation_studies}.

\textbf{Number of local clients:} The results in Figure~\ref{fig:subfig1} shows a trend of increasing performance as we increase the number of local clients. This behavior aligns with expectations and underscores the potential of large-scale federated learning to develop highly performant and generalizable manipulation policies while adhering to strict privacy constraints, such as preventing raw experience sharing between agents.

\textbf{Number of demonstrations per client:} The results in Figure~\ref{fig:subfig2} demonstrate a strong increase in performance as we increase the number of demonstrations available for training. This result suggests that the performance of the centralized policy is heavily influenced by the quality of local imitation learning policies, which improve with larger training datasets~\cite{tuyls2023scaling}.

\textbf{Number of local training epochs:} The results in Figure~\ref{fig:subfig3} reveal an interesting phenomenon: agent performance does not increase monotonically with the number of local training epochs per aggregation round, instead peaking at 25 epochs. We hypothesize that this occurs because local policies overfit to their specific environments after a large number of training epochs, thereby hindering the aggregation process at the central level.

\textbf{Aggregation rounds:} The results in Figure~\ref{fig:subfig4} highlight the need for a continuous evaluation of the performance of the global model: we observe that the success rate of the agent changes significantly across different aggregation rounds. This observation further motivates the need to develop federated learning algorithms specialized for robotic manipulation.


\section{CONCLUSIONS}
\label{sec:conclusions}

This work introduced FLAME, the first federated learning benchmark for robotic manipulation. By leveraging a large-scale dataset collected across diverse simulation environments, FLAME enables the evaluation of federated learning strategies for training manipulation policies in a distributed and privacy-preserving manner. The experimental results demonstrated the feasibility of federated learning for robotic manipulation and highlighted key challenges associated with learning policies in a decentralized framework.

By establishing a standardized benchmark, FLAME lays the groundwork for future research on federated learning in robotic systems. It provides a scalable framework for continual learning and adaptive policy refinement without requiring centralized data aggregation. This benchmark aims to to drive advancements in federated robotics, fostering more robust, generalizable, and privacy-conscious learning methodologies for robotic manipulation tasks.





\section*{ACKNOWLEDGMENT}
This work was supported by the European Research Council (ERC-884807). The computations were enabled by the Berzelius resource provided by the Knut and Alice Wallenberg Foundation at the Swedish National Supercomputer Centre. Additionally, this work was partially supported by the HORIZON-CL4-2021-HUMAN-01 ELSA project.

\bibliographystyle{IEEEtran}
\bibliography{bibliography}

\end{document}